\pdfoutput=1
\documentclass[11pt]{article}

\usepackage{ACL2023}

\usepackage{times}
\usepackage{latexsym}

\usepackage[T1]{fontenc}
\usepackage[utf8]{inputenc}
\usepackage{graphicx}
\usepackage{amsthm}
\usepackage{amsfonts}
\usepackage{amsmath}
\usepackage{amssymb}
\usepackage{inconsolata}
\usepackage{cleveref}

\usepackage{quoting}
\usepackage{todonotes}
\makeatletter
\newcommand*\iftodonotes{\if@todonotes@disabled\expandafter\@secondoftwo\else\expandafter\@firstoftwo\fi}  % defines \iftodonotes{<true>}{<false>}, thanks to \makeatother

\usepackage{microtype}

\usepackage{inconsolata}

\title{Apathetic Conversations}
\title{Stiff Upper Lip}
\title{OK Google, Become Stoic: The Case for Apathy in Conversational AI}
\title{Happiness Bandits}
\title{Conversations as Bland as Honey:\\ How Empathetic Conversational AI Will Ruin Happiness}
\title{Computer says ``No'':\\ The Case Against Empathetic Conversational AI}

\author{Alba Curry \\
  School of Philosophy, Religion \\and History of Science \\
  University of Leeds\\
  \texttt{a.a.cercascurry@leeds.ac.uk} \\\And
  Amanda Cercas Curry \\
  MilaNLP \\
Department of Computing Sciences \\
  Bocconi University \\
  \texttt{amanda.cercas@unibocconi.it} \\}

\begin{document}
\maketitle
\begin{abstract}
%How conversational AI systems respond to a user's emotions is not inconsequential. 
 
Emotions are an integral part of human cognition and they guide not only our understanding of the world but also our actions within it. 
As such, whether we soothe or flame an emotion is not inconsequential.
Recent work in conversational AI has focused on responding empathetically to users, validating and soothing their emotions without a real basis. This AI-aided emotional regulation can have negative consequences for users and society, tending towards a one-noted happiness defined as only the absence of ``negative'' emotions. 
We argue that we must carefully consider whether and how to respond to users' emotions.

\end{abstract}

%\begin{quoting}
%\emph{I would gladly risk feeling bad at times, if it also meant that I could taste my dessert.\\                   \hspace*{\fill}-- Lt. Commander Data (Star Trek)}\quad
%\end{quoting}

\section{Introduction}
Recent work in conversational AI has focused on generating empathetic responses to users' emotional states
\cite[e.g.,][]{ide-kawahara-2022-building,svikhnushina-etal-2022-taxonomy,zhu-etal-2022-multi} as a way to increase or maintain engagement and rapport with the user and to simulate intelligence. 
However, these empathetic responses are problematic.% on several fronts.

First,  while a system might never claim to be human, responses simulating humanness prompt users to further behave as though the systems were \cite{reeves1996media}. 
Empathy, like all emotions, is likely a uniquely human trait %\footnote{We do not exclude the possibility that animals feel emotions such as empathy.} 
and systems that feign it are in effect feigning humanity. 
The ethical issues surrounding anthropomorphism have been discussed at length and are beyond the scope of this paper \citep{salles2020anthropomorphism, bryson2010robots}. 
%\cite{abercrombie-etal-2021-alexa}

Second, empathy requires an ability to both understand and share another's emotions. 
As such, responding empathetically assumes that the system is able to correctly \emph{identify} the emotion, and that it is able to \emph{feel} the emotion itself.\footnote{Correctly identifying an emotion is problematic for animals including human beings. 
However, reasons differ between conversation AI and human beings: Human beings vary in their capacity to identify emotions in part because we struggle at times to identify our own or extend empathy to certain members of society, but we have the capability of identifying emotions. 
Furthermore, our ability to identify the emotions of others builds, at least in part, from our own emotions.}
Neither one of these holds true for conversational AI (or in fact for any AI system).\footnote{Moreover, \citet{barrett2017emotions} have already problematised the identification of human emotions using language or facial expressions in general. }

Third, even if conversational AI were to correctly identify the user's emotions, and perform empathy, we should ethically question the motives and outcomes behind such an enterprise. \citet{svikhnushina-etal-2022-taxonomy} put forward a taxonomy of empathetic questions in social dialogues, paying special attention to the role questions play in regulating the interlocutor's emotions. 
They argue for the crucial role effective question asking plays in successful chatbots due to the fact that often questions are used to express ``empathy'' and attentiveness by the speaker.  
%They distinguish between question acts and question intents. 
Here we highlight the ethical concerns that arise from questions that are characterised by their emotion-regulation functions targeted at the user's emotional state. It is important to note that our argument applies to any use of empathetic AI (see also for example \cite{morris2018towards, de2017simulating}).
%What is supposed to make their dataset stand out above the rest is that others ``do not consider the effect of questions on their addressee's emotional states" (p.2953) 
What happens if the chatbot gets it right? There may be instances where a chatbot correctly identifies that a given situation is worthy of praise and amplifies the pride of the user and the result is morally unproblematic. For example, when \cite{svikhnushina-etal-2022-taxonomy} use the example of amplifying pride in the context of fishing.
%Do the below points A) & B) also relate to the above question? Or is the above question left unanswered? It is not fully clear to me
What happens if it gets it wrong? It depends on the type of mistake:
a) The chatbot fails to put into effect a question's intent, it would be ethically inconsequential; \footnote{In fact, if the chatbot failed to be empathetic then it would simply not engage us in the intended ways.}%\alba{I guess we should keep thinking about whether this is true.}
b) It amplifies or minimises an inappropriate emotion.%\footnote{One may argue that this criticism also applies to human beings, and it does. 
%However, it would be a fallacious argument to insist that just because X does something, it is permissible for Y to do it, too. For what we mean by ``inappropriate emotion'' see section 4.} 
This is the problem we will focus on, arguing that emotional regulation has no place in conversational AI and as such empathetic responses are deeply morally problematic.
While humans will necessarily show empathy for one another, conversational AI cannot understand the emotion and so cannot make an accurate judgement as to its appropriateness. 
This lack of understanding is key as we cannot predict the consequences of moderating an emotion we don't understand, and a dialogue system cannot be held accountable for them.

\section{The Crucial Roles of Emotions}
What emotions are is still up for debate \cite{barrett2017emotions,scarantino2018emotion}. However, their significance for the individual and for society has received renewed interest  \cite{greenspan1995practical,bell2013hard,cherry2021case}.
%There is a repetition of wording for the two below sentences that follow, both use 'important roles', could be the second sentence perhaps rephrased?
Regardless of the emotion model one picks, emotions play important roles, both epistemic and \textit{conative} ones \cite{cercas2022apologia}. %\footnote{For a more detailed discussion see \newcite{cercas2022apologia}. Emotions have more roles, but for the purposes of this paper these are the two that are most useful to emphasise.} 
They perform at least three epistemic roles: (1) They signal to the individual experiencing the emotion what she herself values and how she sees the world (e.g., if you envy your colleague's publications this tells you you value publications and deem yourself similar enough to your colleague that you can compare yourself \cite{protasi2021philosophy}); (2) they signal to others how we see the world; and (3) emotional interactions are invaluable sources of information for third-party observers since they tell us what the members of the interaction value. For example, (1) when you grieve, you signal to yourself and anyone observing that you deem to have lost something of value. It is conceivable that you were unaware up to that point that you valued what you lost---this is captured by the saying ``you don't know what you have till it's gone.'' Furthermore, (2) your friends and family may learn something about you by observing your grief. They too may not have known how much something meant for you. Finally, (3) an observer may also learn about the dynamics of grief (whether it is appropriate to express it for example) by observing whether or not your family validates your grief.
% Perhaps the below sentence could use a conjunctive adverb to make a clear link to the above paragraph

Furthermore, emotions play \textit{conative} roles, meaning that they are involved in important ways with our motivation and desire to act in certain ways.  In other words, not only do some emotions compel and motivate you to act, but also how you act is coloured by the emotion you are experiencing. For example, your anger signals that you perceive that an injustice has occurred. If your boss fails to promote the person who deserves it because of their gender, your anger would motivate you to write a letter of complaint or speak to HR about it.%\footnote{There are good reasons to be sceptical of the claim that we can do this as a result of pure reason, see for example \newcite{brady2013emotional}.} 

Importantly, all emotions, including the so-called ``negative'' emotions (e.g., anger, contempt, hatred, shame, envy, guilt, etc.) %that we have just used as examples 
also share these functions. These emotions are not negative in the sense of being ``bad'', they are called negative because they are painful, and therefore they are emotions that we would tend to avoid for ourselves. A world without injustice would certainly be ideal but we would not want a world of injustice where we were unequipped to notice or become motivated to fight it. Hence why it is imperative that we ask ourselves under which circumstances we ought to enhance or soothe emotions.

\section{The Problem with Empathy}

Literature discussing the value and power of empathy for conversational AI understands empathy as a tool to establish a common ground for meaningful communication and to appear more likeable to users. 
The authors of these studies understand empathy broadly as ``the feeling by which one understands and shares another person's experiences and emotions'' \citep{de2017simulating}.
Empathy facilitates engagement through the development of social relationships, affection, and familiarity. 
Furthermore, for \citet{svikhnushina-etal-2022-taxonomy}, empathy is required in order to enable chatbots to ask questions with emotion regulation intents. 
For example, questions may be used to amplify the user's pride or de-escalate the user's anger, or frustration. 

Empathy, although a common phenomenon, is not a simple one. It enjoys a long history in various scholarly disciplines. Indeed, a lot of ink has been spilled (and still is), for example, over how to make sense of character engagement. How do we, human beings, care for fictional characters? How are we intrigued and moved by their adventures and respond to the emotions and affects expressed in their voices, bodies, and faces as well as imagine the situation they are in and wish them success, closure, or punishment? Empathy is taken to be a key element and yet the exact nature of how human beings are able to experience empathy for fictional characters is currently being debated \citep{tobon2019empathy}. % \amanda{cite chapter "Empathy and Sympathy: Two Contemporary Models of Character Engagement" in The Palgrave Handbook of the Philosophy of Film and Motion Pictures }. 

The reason for highlighting this diversity is that conversational AI would do well to engage seriously with the rich intellectual history of empathy. The definition it tends to engage with lacks the level of complexity required to understand this complex phenomenon. Moreover, it tends to obfuscate the darker sides of empathy. 
%I am a bit confused about what the below sentence is trying to convey. Maybe it would be beneficial to split the sentence into 2 as it is quite long and try to rephrase.
Leaving aside the fact that defining empathy as the ``reactions of one individual to the observed experiences of another'' \citep{de2017simulating} tells us very little about the process by which a human beings, let alone conversational AI, may do this, what we take issue with is what chatbots hope to \textit{do} with that empathy. 
In other words, if for the sake or argument, we presume that conversational AI is able to accurately identify our emotions, the issue of how we deploy empathy is of huge ethical relevance. 

Here we offer a brief summary of three important views against empathy:
\citet{prinz2011against} argues against the common intuition that empathy is by and large a good thing and thus desirable. He raises several issues such as empathy being easily manipulated (such as during a trial), and empathy being partial (we are more empathetic towards people we perceive to be of our own race, for example). Both claims have been empirically verified. Thinking about how this might affect empathetic conversational AI for example in the case of using them for social assistive robots, we might worry if based on its empathetic reactions it chose to help certain people over others. 

Taking the argument further, \citet{bloom2017against} argues against empathy and for what he calls rational compassion. He contends that empathy is one of the leading motivators of inequality and immorality in society. Thus, far from helping us to improve the lives of others, empathy is a capricious and irrational emotion that appeals to our narrow prejudices; it muddles our judgement and, ironically, often leads to cruelty. Instead, we ought %to not rely on empathy, but 
to draw upon a more distanced compassion.\footnote{Assessing Bloom's argument with regards to rational compassion and whether it would be feasible for conversational AI is beyond the scope of this paper although worthy of pursuit.} 

% Leave out either 'Our' or 'the' from the sentence following (1) 
There are three lessons we can take from this: (1) Given empathy's prejudices, we would need to think deeper about how to mitigate them in conversational AI; (2) Given that empathy is used not just know what brings people pleasure, but also what brings pain, we might want to question the general future uses of empathy in conversational AI; (3) if we buy Bloom's argument, then conversational AI should consider not imitating human beings, but becoming agents of rational compassion.

\citet{breithaupt2019dark} also takes issue with empathy, arguing that we commit atrocities not out of a failure of empathy, but rather as a direct consequence of successful, even overly successful, empathy. He starts the book by reminding us that ``[e]xtreme acts of cruelty require a high level of empathy.''

The further lesson we can take from this is that while people generally assume that empathy leads to morally correct behaviour, and certainly there are many positive sides of empathy, we should not rely on an overly simple or glorified image of empathy.

However, our problem is not necessarily with empathy per se, but rather with the explicit functions conversational AI hopes to achieve with it, namely to enhance engagement, to inflate emotions deemed positive, and to soothe emotions deemed negative (e.g., \citeauthor{svikhnushina-etal-2022-taxonomy}, \citeyear{svikhnushina-etal-2022-taxonomy}). 
Our claim is that we ought to think carefully about the consequences of soothing negative emotions only because they we have a bias against them. 
Not only is this approach based on a naive understanding of emotions, it fails to recognise the importance of human beings being allowed to experience and express the full spectrum of emotions. 
One ought to not experience negative emotions because there is nothing to be upset about, not because we have devised an emotional pacifier. 
In other words, the issue is that conversational AI lacks a sound value system for deciding why certain emotions are validated and others soothed. Furthermore, this AI-aided emotional regulation can have negative consequences for users and society, tending towards a one-noted notion of happiness defined as only the absence of ``negative'' emotions.

\section{When Emotions Get Things Wrong}

There are two illustrative problems with the kinds of decisions behind amplifying and de-escalating emotions. 
One is the problem of what the ideal character might be. When you talk to a friend they will decide whether to soothe or amplify your emotions based not just on the situation but also on who they deem you to be. If they think you are someone who has a hard time standing up for yourself they will amplify your anger to encourage you to fight for yourself, but if they think you are someone who leans too much on arrogance, they will de-escalate your sense of pride---even if, all things being equal, your pride on that occasion was warranted. Hence, not only would a conversational AI require prior knowledge of the interlocutor in terms of her character, but furthermore it would have to decide what are desirable character traits.

The second question regards what an ideal emotion in a particular situation might be. We may all find it easy to say that negative emotions such as anger often get things wrong and lead to undesirable outcomes. However, positive emotions such as joy, hope, or pride which we may intuitively wish to amplify can also get things wrong.
We assess and criticise emotions along a number of distinct dimensions:  Firstly, emotions  may  be  criticised  when  they  do  not  fit  their  targets. You may, for example, be open to criticism for feeling fear in the absence of danger. 
Unfitting emotions fail to correctly present the world.
In the case of pride, would we want to amplify someone's pride if they either did not in fact achieve anything, or if their achievement was not merited? For example, if their nephew did very well in maths when in fact we know their nephew cheated?
Second, an emotion may be open to criticism when it is not based on good evidence or is unreasonable.  Consider  the  person  who  suffers  from  hydrophobia:  Given  that  in  the  vast majority of situations water is not dangerous, this person’s fear is both unreasonable and unfitting.  
But  even  fitting  emotions  may  be  unreasonable.  One  may,  for  example,  be terrified of tsunamis because one believes that they cause genetic mutations. 
In this case, one’s fear is fitting---tsunamis are very dangerous---yet the fear is unreasonable since it is not based on good reasons. 
Third, an emotion may be criticised because it isn’t prudent to feel. 
We might warn someone not to show anger when interacting with a person with a gun since they might get themselves killed; anger in this case may be reasonable and fitting given the gunman's actions and yet imprudent. 
Finally, we may condemn emotions as morally non-valuable  because  of  the  unacceptable  way  in  which  they  present  their  targets, e.g.,   
one may, argue  that \textit{schadenfreude}  is  morally  objectionable  because  it  presents the  pain  of  another  person  as laughable.

Positive emotions may be unfitting, unreasonable, and imprudent, as well as morally condemnable just as negative emotions may well be fitting, reasonable, and prudent, as well as morally laudable. 
In other words, even if one is equipped with empathy there are crucial normative decisions involved in question intents aimed at emotional regulation.\footnote{See the complex example in \newcite{silva2021epistemic}} 
Amplifying and de-escalating emotion inappropriately %, as in the case of what is best for one according to one's character and situation, as well as amplifying and de-escalating the wrong emotions 
can have devastating moral outcomes. 

%\section{The Problem of Anthropomorphism}

%the problem of authority
%Anthropomorphism refers to a system's humanlikeness, whether in terms of its embodiment or its behaviour. When it comes to behaviour, there are several aspects that contribute to how anthropomorphic a system is: whether it is interactive, whether it fulfils a human role.Dialogue systems, regardless of their embodiment, are inherently anthropomorphic because of their use of language and their interactiveness. In addition, these systems are often given personas that further increase how human-like they are through the use of human-specific traits like favourite foods, colours and comedians \cite{abercrombie-etal-2021-alexa}. Anthropomorphism and personification help users feel more connected to a system as they ascribe humanity to it.The EU and the U.S. state of California have already put forth regulation against systems claiming to be human and recent work has focused on ensuring systems can confirm their non-human identity \citep{gros-etal-2021-r}. Possibly itneresting paper: \cite{torres-fonsesca-kennington-2022-hadreb}

\section{Empathy and Responsibility}

%? Why are empathetic responses bad when they come from AI and not from humans

Human beings, all things being equal, will inevitably experience empathy. A reasonable human being experiencing empathy for another is proof of the importance of someone else's emotional state---for better or for worse. 
This supports the idea that our emotions are important, as opposed to the notion that they hinder rationality and ought to be regulated.
They tell us many things about our world. 

Similarly to many NLP systems' understanding of language, the empathetic responses of conversational AI are only performative \cite{bender2020climbing}. 
Thus, they provide a false sense of validity or importance. 
What if someone is experiencing an unfitting, unreasonable, or morally reprehensible emotion? Should a chatbot still showcase empathy? We hope to have shown that such decisions are deeply morally problematic and complex. 
 
Hence, another key problem is responsibility. A human agent may choose to express their empathy (even if they cannot choose feeling it) and they may choose to attempt to regulate someone else's emotions based on their knowledge of the situation and the speaker's character. If a human being wrongly regulates someone else's emotions, they will be morally responsible for the consequences. Who is morally responsible in the case of conversational AI agents? Who are they benefiting when they are not actually benefiting the human agent? This issue is further elaborated on by \citet{veliz2021moral}. %\amanda{Moral zombies? \citep{veliz2021moral}}

\section{Related Work}
Our article sits at the intersection of emotion detection, response generation, and safety in conversational AI. We keep this section brief as we cite relevant work throughout the article.
Several works have already focused on the issue of giving AI systems sentience, such as \citet{bryson2010robots}. While this could make the systems truly empathetic, we agree that we have a duty not to create sentient machines. 

\citet{lahnala2022critical} problematise NLP's conceptualisation of empathy which, they argue, is poorly defined, leading to issues of data validity and missed opportunities for research. Instead, we argue that even a more specific definition of empathy presents ethical issues that cannot be overlooked or ignored and must carefully evaluated.

\citet{dinan-etal-2022-safetykit} provide a framework to classify and detect safety issues in end-to-end conversational systems. 
In particular, they point out systems that respond inappropriately to offensive content and safety-critical issues such as medical and emergency situations.
We could apply their framework to empathetic responses where the system takes the role of an ``impostor'': empathetic responses require a system to pretend to understand the emotion. However, the extent to which emotions play a role in human cognition and what the consequences of regulating these emotions for the users are has not been discussed in the literature to the best of our knowledge.

\section{Conclusion}
In this position paper, we argued that emotional regulation has no place in conversational AI and as such empathetic responses are deeply morally problematic.
While humans will necessarily show empathy for one another, conversational AI cannot understand the emotion and so cannot make an accurate judgement as to its reasonableness. 
This lack of understanding is key because we cannot predict the consequences of assuaging or aggravating an emotion, and a dialogue system cannot be held accountable for them. 
We hope to encourage reflection from future researchers and to initiate a discussion of the issue, not only in this particular case but also more reflection when it comes to pursuing seemingly positive goals such as bringing disagreeing parties towards agreement. Like with other ethically sensible topics, the community should come together to agree on a strategy that minimises harm.

\section*{Limitations}
While we strongly argue against empathetic conversational systems, there may be use cases -- such as psychotherapy or educational chatbots -- where validating a user's emotions is, if not required, helpful in terms of their goal.  In addition, while a lot of the work on empathetic responses we have discussed is intentional, generative models like ChatGPT produce relatively uncontrolled responses that may well be unintentionally empathetic. As with toxic outputs, care should be taken to prevent these models from validating users' emotions that cannot be understood. 

\section*{Acknowledgements}
We thank the anonymous reviewers and Gavin Abercrombie for their thorough and helpful comments. 
This project has partially received funding from the European Research Council (ERC) under the European Union’s Horizon 2020 research and innovation program (grant agreement No.\ 949944, INTEGRATOR).
Amanda Cercas Curry is a member of the MilaNLP group and the Data and Marketing Insights Unit of the Bocconi Institute for Data Science and Analysis.

\bibliography{anthology}
\bibliographystyle{acl_natbib}

\end{document}